\documentclass[accepted]{uai2021} 
\usepackage[british]{babel}
\usepackage{natbib}
\usepackage{mathtools,bm}
\usepackage{booktabs}
\usepackage{tikz} 
\usepackage{amsmath}
\usepackage{xspace}
\usepackage[RPvoltages]{circuitikz}
\usepackage{pgfplots,enumerate}
\pgfplotsset{compat=1.7}
\usepackage{graphicx}
\usepackage{booktabs}
\usepackage{algorithm}
\usepackage{algorithmic}
\usepackage{caption}
\usepackage{subcaption}
\usepackage{amssymb}
\usepackage{subcaption}

\newtheorem{prp}{Proposition}

\newcommand{\interp}[1]{\langle #1 \rangle}
\newcommand{\slopp}{\texttt{SLoPP}}

\title{Structural Learning of Probabilistic Sentential Decision Diagrams\\under Partial Closed-World Assumption}
\author{\href{mailto:<alessandro@idsia.ch}{Alessandro Antonucci}}
\author{\href{mailto:<alessandro.facchini@idsia.ch}{Alessandro Facchini}\thanks{Supported by the Hasler foundation grant n. 20061.}}
\author{\href{mailto:<lilith@idsia.ch}{Lilith Mattei}}
\affil{%
Istituto Dalle Molle di Studi per l'Intelligenza Artificiale (IDSIA)\\
Lugano, Switzerland}
\begin{document}
\maketitle
\begin{abstract}
Probabilistic sentential decision diagrams are a class of structured-decomposable probabilistic circuits especially designed to embed logical constraints. To adapt the classical \textsc{LearnSPN} scheme to learn the structure of these models, we propose a new scheme based on a partial closed-world assumption: data implicitly provide the logical base of the circuit. Sum nodes are thus learned by recursively clustering batches in the initial data base, while the partitioning of the variables obeys a given input vtree. Preliminary experiments show that the proposed approach might properly fit training data, and generalize well to test data, provided that these remain consistent with the underlying logical  base, that is a relaxation of the  training data base. 
\end{abstract}
\section{Introduction}\label{sec:intro}
Probabilistic Circuits (PCs) have been recently introduced as a general computational framework unifying the different existing formalisms for tractable probabilistic modeling \citep{choiprobabilistic}.

Learning the structure of a PC from data is emerging as a crucial challenge for a widespread application of PCs in the area of machine learning. Different algorithms have been (e.g., \cite{lowd2012learning}) and still are (e.g., \cite{peharz2020einsum}) designed to improve the state of the art in this field. Most of these algorithms are somehow inspired by the \textsc{LearnSPN} scheme of \cite{gens2013learning}, the first algorithm for the structural learning of general PCs, originally designed for sum-product networks \citep{poon2011sum}, the most popoular class of PCs.

In this position paper, we focus on the problem of learning from 
data, without other prior domain knowledge, the structure of a particular class of PCs, called \emph{Probabilistic Sentential Decision Diagrams} (PSDDs, \cite{kisa2014probabilistic}. Roughly speaking, a PSDD is a weighted logical circuit inducing a probability distribution assigning non-zero probability only to states consistent with the Boolean formula encoded by the circuit. 

As a matter of fact there are only two main algorithms to learn the structure of a PSDD: \textsc{LearnPSDD} \citep{liang2017learning} and \textsc{Strudel} \citep{DangPGM20}. Despite their differences, both these algorithms start from a circuit subject to local transformations. These local actions are designed to preserve the underlying logical base of the circuit, which should therefore be specified when the initial circuit is 
given, meaning that  both algorithms may assume some prior domain knowledge modelled by the input circuit. 

The \textsc{LearnSPN} architecture is different: the algorithm learns a circuit from an input database by growing it from the root, rather than performing local operations on an initial circuit. \textsc{LearnSPN} takes as input a database and recursively divides it into sub-databases by finding almost independent subsets of variables and clustering similar instances. 
 
Two structural properties of PSDDs prevent a straightforward application of \textsc{LearnSPN}. First, PSDDs are structured-decomposable, this roughly meaning that the splitting of the variables into independent sets must agree to some constraint. Second, a PSDD encodes a Boolean formula representing a knowledge base (KB); thus it is desirable to steer the operations of \textsc{LearnSPN} so that the obtained KB represents the input database. Here we address such situation by making \textsc{LearnSPN} able to learn a PSDD encoding a KB describing the data. 

Of course there is no unique way to agree on what ``describing the data'' means. In database theory, a traditional assumption, called \emph{closed-world assumption} (CWA), states that a database reflects with fidelity  the concerned state of affairs, that it is complete \citep{reiter1981closed}. Stated otherwise, a fact that is true in the world is stored in the database, whereas a fact that is missing in the database just does not hold in the world. 
This assumption is usually followed in ILP (Inductive Logic Programming), a branch of Machine Learning whose objective is to induce a set hypothesis represented as logic programs (rules). The opposite of the closed-world assumption is the \emph{open-world assumption} \citep{reiter1981closed,imielinski1989incomplete}. Commonly employed in the area of description logic \citep{baader2003description}, this simply states that some facts holding in the real world may be missing, and thus does not presume that the knowledge of a domain is complete. 

However, in many cases, it may be desirable to have an intermediate perspective: some parts of the observed data may be complete, and some others may be incomplete. First treated by \cite{motro1989integrity}, this view has later been instantiated in various forms under various names (local closed-world assumption, partial closed-world assumption, partial completeness, and others; see, e.g., \cite{galarraga2013amie,darari2013completeness,dong2014knowledge})

In this work we follow a similar intermediate perspective. More precisely, we extract a KB from a database $\mathcal{D}$ by recursively applying an instance of a \emph{partial closed-world assumption} (PCWA). That is, each step of the procedure extracts a sub-database $\mathcal{D}'$ and a subset $\bm{X}'$ of variables for which $\mathcal{D}'$ is assumed to reflect completely the concerned possible states of affairs (given the constraints provided by the previous step). This idea is made precise in Section \ref{sec:growing}. 

The paper is organised as follows. In Section \ref{sec:background} we provide the necessary background about PSDDs. Our contribution is in Section \ref{sec:growing}, where we present a PCWA based structural learning algorithm for PSDDs inspired by the \textsc{LearnSPN} schema. In Section \ref{sec:ex} we run preliminary experiments. Conclusions and outlooks are in Section \ref{sec:conc}.

\section{Background}\label{sec:background}
\subsection{Basics}\label{subsec:notation}
We focus on a set of Boolean variables $\bm{X}:= (X_1,\ldots,X_n)$. Let $\mathcal{D}$ denote a database of joint states of $\bm{X}$, to be also called \emph{records} and denoted as $\bm{x} = (x_1,\ldots, x_n)$. The database obtained by taking only the columns of $\mathcal{D}$ relative to variables in $\bm{X}'\subseteq \bm{X}$ is denoted as $\mathcal{D}^{\bm{X}'}$. When coping with a single variable $X\in\bm{X}$, with a small abuse of notation, we intend as $\mathcal{D}^{X=1}$ the (sub)set of true instances of $\mathcal{D}^{X}$, and we analogously intend $\mathcal{D}^{X=0}$. Notation $|\mathcal{D}|$ is used instead to denote the number of records in $\mathcal{D}$. If the records are indexed by the integers $\{1,\ldots, m\}$, with $m:=|\mathcal{D}|$, we denote as $\mathcal{D}_J$ the database obtained by taking only the records indexed by the elements of $J$, with $J \subseteq \{1,\ldots,m\}$.
Let $ \bm{x}^{j}=x_1^{j},\ldots,x_n^{j}$ denote the $j$-th record of $\mathcal{D}$. According to CWA, we assume the database $\mathcal{D}=\{ \bm{x}^1, \dots, \bm{x}^m\}$ complete, this meaning that it completely describes the possible state of (the concerned aspects of) the world. Hence, $\mathcal{D}$ induces a Disjunctive Normal Form (DNF) 
\begin{equation}
\phi(\mathcal{D}) := \mathop{\bigvee}_{j=1}^m \bigwedge_{i=1}^n L_i^{j} \,,
\end{equation}
whose conjunctive clauses are the ones describing the records belonging to $\mathcal{D}$. That is, for each record $\bm{x}^j:=(x^j_1, \dots, x^j_n) \in \mathcal{D}$, the corresponding conjunctive clause under the CWA coincides with  $L^j_1 \wedge \dots \wedge L^j_n$ where, for $i\leq n$ 
\begin{equation}
L^j_i=\begin{cases}
X_i & \text{when }x^j_i=1\,,\\
\neg X_i & \text{otherwise.}
\end{cases}
\end{equation}

\subsection{PSDDs}\label{subsec:psdd}
Inspired by \cite{choiprobabilistic}, in this section we introduce PSDDs as a class of PCs that, in addition to specific structural properties (e.g., determinism and structured-decomposability), encode Boolean formulae.

\subsubsection{Probabilistic Circuits}
A PC $\mathcal{C}$ over $\bm{X}$ is a rooted DAG $\mathcal{G}$ annotated by parameters $\bm{\theta}$ in order to represent a joint probability mass function $\mathbb{P}(\bm{X})$. Formally, each leaf (terminal node) $g$ of $\mathcal{G}$ is called an \emph{input unit} and it is associated with a variable $X_g\in\bm{X}$ and a (possibly degenerate) probability mass function $d_g(X_g)$. 
Non-leaves nodes can be \emph{sum units} and \emph{product units}. Parameters are associated with both input units and sum units. In case of a sum unit, these can be seen as the probabilities of a mass function weighting its outgoing arcs, while for input units a mass function $\pi$ is directly provided. In practice, if ${\sf ch}(g)$ are the children of a node $g$ (also called its \emph{inputs}) and $\bm{X}_g$ the set of variables associated to the input units reachable from $g$, and called its \emph{scope},  the corresponding probability mass function $\mathbb{P}_g(\bm{X}_g)$ is defined as follows:
\[
\mathbb{P}_g(\bm{x}_g):=
\begin{cases}
\pi_g(\bm{x}_g) & \text{if $g$ is an input unit,}\\
\prod_{\ell \in {\sf ch}(g)}\mathbb{P}_\ell(\bm{x}_\ell) & \text{if $g$ is a product unit,}\\
\sum_{\ell \in {\sf ch}(g)} \theta_{\ell,g}\mathbb{P}_\ell(\bm{x}_\ell) & \text{if $g$ is a sum unit.}
\end{cases}
\]
As we focus here on Boolean variables, we can identify input units with literals or logical constants, sum units with disjunctions, and product units with conjunctions. Graph $\mathcal{G}$ can be consequently seen as a logical circuit encoding a Boolean formula. Both logical and probabilistic elements are considered by the PSDD definition provided in the next section.

\subsubsection{Defining PSDDs}
PSDDs have been originally presented as a probabilistic extension of a class of logical circuits called \emph{sentential decision diagrams} \citep{darwiche2011sdd}. The extension is obtained by respecting the formula in the underlying circuit. Thus, unlike PCs, the position of the variables in a PSDD is constrained by a meta-structure called \emph{vtree}.

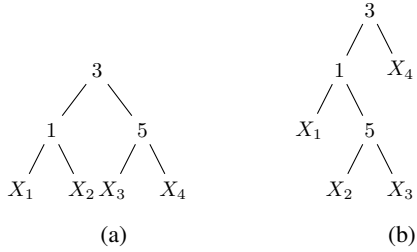
\begin{figure}[htp!]
\centering
\begin{subfigure}[b]{3cm}
\begin{tikzpicture}[scale=0.8,transform shape,level distance=1cm,level 1/.style={sibling distance=1.5cm},level 2/.style={sibling distance=1cm}]
\node {$3$}child {node {$1$} child {node {$X_1$}}
child {node {$X_2$}}} child {node {$5$}
child {node {$X_3$}} child {node {$X_4$}}};
\end{tikzpicture}
\caption{\label{fig:vtree1}}
\end{subfigure}
\quad\quad
\begin{subfigure}[b]{3cm}
\begin{tikzpicture}[scale=0.8,transform shape,level distance=1cm,
level 1/.style={sibling distance=1cm},level 2/.style={sibling distance=1cm}]
\node {$3$} child {node {$1$} child {node {$X_1$}} child { node {$5$} child {node {$X_2$}} child {node {$X_3$}}}} child {node {$X_4$}};
\end{tikzpicture}
\caption{\label{fig:vtree2}}
\end{subfigure}
\caption{Two vtrees over $\bm{X}=\{X_1,X_2,X_3,X_4\}$.}\label{fig:vtree}
\end{figure}

A vtree over $\bm{X}$ is just a full binary tree whose leaves are in one-to-one correspondence with the variables in $\bm{X}$ (see, e.g., Figure \ref{fig:vtree}). We intend vtrees as recursive objects and denote them with their root node. In practice, each node $u$ of vtree $v$ is a(sub-)vtree. Notation $\bm{v}$ is used for the variables of $v$. Moreover, we denote as $v^l$ and $v^r$ the sub-vtrees of $v$ rooted at its left and right children.

Using PCs terminology, we can define a PSDD associated to a vtree $v$ by the properties of its structure and its parameters.
\paragraph{PSDD Structure.}
PSDD graph $\mathcal{G}$ is defined as follows.
\begin{itemize}
\item Input units are associated with the leaves of vtree $v$, i.e., have scope $\{X\}$ for some  $X\in \bm{v}$. They can be literals ($X$ or $\neg X$) or the logical constant $\top$ (\emph{true}). 
\item A product unit $n$ has two inputs. Notation $n=(p,s)$ is used, $p$ is called its \emph{prime} and $s$ its \emph{sub}.
\item A sum unit $n$ is associated with an internal node $u$ of $v$, its inputs are product units and are called its \emph{elements}. The notation  $n:=\{(p_i,s_i)\}_{i=1}^k$ is used. Its primes and subs are either sum or input units. The scope of $n$ and of its elements is $\bm{u}$. The scope of the primes is $\bm{u^l}$ and the scope of the subs is $\bm{u^r}$.
\item The root is a sum unit. 
\end{itemize}

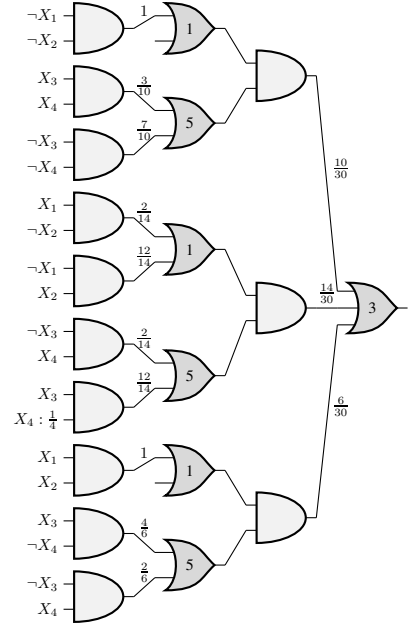
\begin{figure}[htp!]
\centering
\begin{tikzpicture}[scale=0.6,transform shape]
\draw (0,0) node[and port,fill=black!5] (myand1){};
\draw (0,-1.4) node[and port,fill=black!5] (myand2){};
\draw (0,-2.8) node[and port,fill=black!5] (myand3){};
\draw (0,-4.2) node[and port,fill=black!5] (myand4){};
\draw (0,-5.6) node[and port,fill=black!5] (myand5){};
\draw (0,-7.0) node[and port,fill=black!5] (myand6){};
\draw (0,-8.4) node[and port,fill=black!5] (myand7){};
\draw (0,-9.8) node[and port,fill=black!5] (myand8){};
\draw (0,-11.2) node[and port,fill=black!5] (myand9){};
\draw (0,-12.6) node[and port,fill=black!5] (myand10){};
\draw(2,0) node[or port,fill=black!15] (myor1){1};
\draw(2,-2.1) node[or port,fill=black!15] (myor2){5};
\draw(2,-4.9) node[or port,fill=black!15] (myor3){1};
\draw(2,-7.7) node[or port,fill=black!15] (myor4){5};
\draw(2,-9.8) node[or port,fill=black!15] (myor5){1};
\draw(2,-11.9) node[or port,fill=black!15] (myor6){5};
\draw(4,-1.05) node[and port,fill=black!5] (myand1b){};
\draw(4,-6.2) node[and port,fill=black!5] (myand2b){};
\draw(4,-10.85) node[and port,fill=black!5] (myand3b){};
\draw(6,-6.2) node[or port,fill=black!15,number inputs=3] (root){3};
\draw (myand1.in 1) node[anchor=east] {$\neg X_1$};
\draw (myand1.in 2) node[anchor=east] {$\neg X_2$};
\draw (myand2.in 1) node[anchor=east] {$X_3$};
\draw (myand2.in 2) node[anchor=east] {$X_4$};
\draw (myand3.in 1) node[anchor=east] {$\neg X_3$};
\draw (myand3.in 2) node[anchor=east] {$\neg X_4$};
\draw (myand4.in 1) node[anchor=east] {$X_1$};
\draw (myand4.in 2) node[anchor=east] {$\neg X_2$};
\draw (myand5.in 1) node[anchor=east] {$\neg X_1$};
\draw (myand5.in 2) node[anchor=east] {$X_2$};
\draw (myand6.in 1) node[anchor=east] {$\neg X_3$};
\draw (myand6.in 2) node[anchor=east] {$X_4$};
\draw (myand7.in 1) node[anchor=east] {$X_3$};
\draw (myand7.in 2) node[anchor=east] {$X_4:\frac{1}{4}$};
\draw (myand8.in 1) node[anchor=east] {$X_1$};
\draw (myand8.in 2) node[anchor=east] {$X_2$};
\draw (myand9.in 1) node[anchor=east] {$X_3$};
\draw (myand9.in 2) node[anchor=east] {$\neg X_4$};
\draw (myand10.in 1) node[anchor=east] {$\neg X_3$};
\draw (myand10.in 2) node[anchor=east] {$X_4$};
\draw (myand1.out) -- (myor1.in 1) node[midway,above]{$1$};
\draw (myand2.out) -- (myor2.in 1) node[midway,above]{$\tfrac{3}{10}$};
\draw (myand3.out) -- (myor2.in 2) node[midway,above]{$\tfrac{7}{10}$};
\draw (myand4.out) -- (myor3.in 1) node[midway,above]{$\tfrac{2}{14}$};
\draw (myand5.out) -- (myor3.in 2) node[midway,above]{$\tfrac{12}{14}$};
\draw (myand6.out) -- (myor4.in 1) node[midway,above]{$\tfrac{2}{14}$};
\draw (myand7.out) -- (myor4.in 2) node[midway,above]{$\tfrac{12}{14}$};
\draw (myand8.out) -- (myor5.in 1) node[midway,above]{$1$};
\draw (myand9.out) -- (myor6.in 1) node[midway,above]{$\tfrac{4}{6}$};
\draw (myand10.out) -- (myor6.in 2) node[midway,above]{$\tfrac{2}{6}$};
\draw (myor1.out) -- (myand1b.in 1) node[]{};
\draw (myor2.out) -- (myand1b.in 2) node[]{};
\draw (myor3.out) -- (myand2b.in 1) node[]{};
\draw (myor4.out) -- (myand2b.in 2) node[]{};
\draw (myor5.out) -- (myand3b.in 1) node[]{};
\draw (myor6.out) -- (myand3b.in 2) node[]{};
\draw (myand1b.out) -- (root.in 1) node[midway,above right]{$\tfrac{10}{30}$};
\draw (myand2b.out) -- (root.in 2) node[midway,above]{$\tfrac{14}{30}$};
\draw (myand3b.out) -- (root.in 3) node[midway,above right]{$\tfrac{6}{30}$};
\end{tikzpicture}
\caption{A PSDD over four variables trained from the database in Table \ref{tab:data}. Its vtree is in Figure \ref{fig:vtree1}.}
\label{fig:psdd}
\end{figure}

\paragraph{PSDD Parameters.} Parameters $\bm{\theta}$ are defined as follows.
\begin{itemize}
\item Each input unit $\top$ associated with variable $X$ is annotated with a weight $\theta \in ]0,1[$.
\item Each sum unit $\{(p_i,s_i)\}_{i=1}^k$ is annotated with non-negative weights $\theta_1, \dots, \theta_k$ such that $\mathop{\Sigma}_{i=1}^k \theta_i=1$.
\end{itemize}
\paragraph{PSDD Logical Base} Each PSDD node $n$ encodes a Boolean formula $\interp{n}$ called its (logical) \emph{ base} and defined recursively as follows.
\begin{itemize}
\item For input units:  $\interp{X}:= X$, $\interp{\neg X}:=\neg X$, and $\interp{\top}:=\top$
\item If $n= (p,s)$ is a product unit, $\interp{n}:=\interp{p}\wedge \interp{s}$
\item If $n$ is a sum unit, $n=\{n_i\}_{i=1}^k$,
$\interp{n}:=\mathop{\bigvee}_{i=1}^k \interp{n_i}$.
\end{itemize}
In fact, product units represent AND gates and sum units represent OR gates. The primes of a sum unit are defined as exclusive, i.e., $\interp{p_i} \wedge \interp{p_j} \equiv \bot$ for $i\neq j$.

\paragraph{Induced Probability Mass Function.}
Each unit $n$ induces a probability mass function $\mathbb{P}_n$ over the variables of its scope.
\begin{itemize}
\item If $n$ is an input unit  with scope $\{X\}$, $\mathbb{P}_n$ is a univariate mass function: $\mathbb{P}_{\top}(X):=(\theta, 1-\theta)$;
$\mathbb{P}_{X}(X):=(1,0)$; and
$\mathbb{P}_{\neg X}(X):=(0,1)$.
\item Otherwise, let $(\bm{X},\bm{Y})$ be the scope of $n$, where $\bm{X}$ are its left variables and $\bm{Y}$ its right ones.
\begin{itemize}
\item If $n=(p,s)$ is a product unit,
\begin{equation}
\mathbb{P}_n(\bm{x},\bm{y}):=  \mathbb{P}_{p}(\bm{x})\mathbb{P}_{s}(\bm{y})\,.
\end{equation}
\item If $n=\{n_i\}_{i=1}^k$ is a sum unit, 
\begin{equation}
\mathbb{P}_n(\bm{x},\bm{y}):= \theta_i \mathbb{P}_{n_i}(\bm{x},\bm{y}) \,,
\end{equation}
for the unique $1\leq i \leq k$ such that $\bm{x}\models\interp{p_i}$. Moreover, each input of a sum unit assigns non-zero probability to at least one world.\footnote{This means that, unlike in the original definition, and as done in \citep{liang2017learning}, we do not consider elements encoding the false, aka \emph{dead branches}.}
\end{itemize}
\end{itemize}
Let us note that the exclusivity of the primes makes PSDDs a special class of \emph{deterministic} PCs. Indeed, for each decision node $n = \{(p_i,s_i)\}_{i=1}^k$ and for each world $(\bm{x}\bm{y})$, there is at most one prime $p_i$ such that $\mathbb{P}_{p_i}(\bm{x}) > 0$. 
Moreover, the induced probability mass function of a PSDD factorizes through its variables by following its vtree. This makes PSDDs structured-decomposable PCs.

As a demonstrative example, Figure \ref{fig:psdd} depicts a PSDD over four variables consistent with the vtree in Figure \ref{fig:vtree}.a and whose parameters have been trained from the database in Table \ref{tab:data} as done in \citep{kisa2014probabilistic}. Labels on the decision nodes denote the vtree nodes for which the sub-PSDD is associated.

\begin{table}[htp!]
\centering
\begin{tabular}{ccccc}
\toprule
\#&$X_1$&$X_2$&$X_3$&$X_4$\\
\midrule
$3$&0&0&1&1\\
$7$&0&0&0&0\\
\hline
$2$&1&0&0&1\\
$3$&0&1&1&1\\
$9$&0&1&1&0\\
\hline
$2$&1&1&0&1\\
$4$&1&1&1&0\\
\hline
0&1&0&1&1\\
0&1&0&1&0\\
0&0&1&0&1\\
\bottomrule
\end{tabular}
\caption{A database for the joint states of four Boolean variables. The last three records are unobserved.}
\label{tab:data}
\end{table}

\subsection{\textsc{LearnSPN}}\label{sec:learnspn} 
To conclude the review of the background material, let us consider the pseudo-code in Algorithm \ref{alg:learnspn} where an adaptation of the classical \textsc{LearnSPN} scheme for the learning of PCs from a database is depicted \citep{gens2013learning}. This is a recursive procedure, where the database subject to horizontal (line 4) and vertical (line 6) splits. By horizontal split we intend a partition in clusters of similar records based on a clustering algorithm generically denoted as \textsc{Cluster} and returning the index sets of the different clusters. A vertical split is intended instead a partition of the variables in the database achieved by some independence test and denoted as \textsc{Partition}. When single columns are found (line 1), univariate mass functions are directly obtained from the data. The recursion is achieved by expressing the output mass function as a weighted sum of the mass functions associated to the different clusters (sums in line 8) and expressing them as a product of the mass functions associated with the different groups of variables (products in line 8).

\begin{algorithm}[htb!]
\caption{${\textsc{LearnSPN}}(\mathcal{D},\bm{X})$ (adaptated from \cite{gens2013learning})\\Return a PC given database $\mathcal{D}$ over $\bm{X}$}\label{alg:learnspn}
\begin{algorithmic}[1] 
\IF{$|\bm{X}|=1$}
\RETURN $\pi(\bm{X}) \leftarrow \mathcal{D}$
\ELSE
\STATE $\{J_i\}_{i=1}^k \leftarrow \textsc{Cluster}(\mathcal{D}^{\bm{X}})$
\FOR{$i\leftarrow 1,\ldots,k$}
\STATE $\{\bm{X}^{j}_i\}_{j=1}^{m_i}
\leftarrow {\textsc{Partition}}(\bm{X},\mathcal{D}_{J_i})$
\ENDFOR
\RETURN 
$\sum_{i=1}^k 
\frac{|\mathcal{D}_{J_i}|}{|\mathcal{D}|}
\prod_{j=1}^{m_i} {\textsc{LearnSPN}}
(\mathcal{D}_{J_i}^{\bm{X}_j^i},\bm{X}_j^i)$
\ENDIF
\end{algorithmic}
\end{algorithm}

\section{Learning PSDDs under PCWA}\label{sec:growing}
The goal of this section is to show how the \textsc{LearnSPN} scheme as sketched in Section \ref{sec:learnspn} can be modified in order to force the output of Algorithm \ref{alg:learnspn} to be a valid PSDD. Remarkably, this will be shown to correspond to a PCWA approach. Similarly to \textsc{LearnSPN}, we might obtain a PSDD by recursively perform horizontal and vertical splits of $\mathcal{D}$. Yet, unlike \textsc{LearnSPN}, the horizontal splits should be driven by the vtree, which we assume here available as an input.

\begin{algorithm}[htb!]
\caption{$\textsc{SLoPP}(\mathcal{D},v)$\\Return a PSDD given database $\mathcal{D}$ and vtree $v$ and  over $\bm{X}$}\label{alg:slopp}
\begin{algorithmic}[1] 
\IF{$|\bm{v}|=1$}
\STATE $X\leftarrow$ unique variable in  $\bm{v}$
\IF{$|\mathcal{D}^{X=1}|=|\mathcal{D}|$}
\RETURN $X$
\ELSIF{$|\mathcal{D}^{X=0}|=|\mathcal{D}|$}
\RETURN $\neg X$
\ELSE
\RETURN $(\top, \tfrac{|\mathcal{D}^{X=1}|}{|\mathcal{D}|})$
\ENDIF
\ELSE
\STATE $\{J_i\}_{i=1}^k \leftarrow \textsc{Cluster}(\mathcal{D}^{\bm{v}^l})$
\FOR{$i=1, \dots, k$}
\STATE $p_i \leftarrow {\textsc{SLoPP}}(\mathcal{D}^{\bm{v}^l}_{J_i}, v^l)$
\STATE $s_i \leftarrow {\textsc{SLoPP}}(\mathcal{D}^{\bm{v}^r}_{J_i}, v^r)$
\ENDFOR
\RETURN $\{(p_i,s_i,\tfrac{|\mathcal{D}_{J_i}|}{|\mathcal{D}|})\}_{i=1}^k$
\ENDIF
\end{algorithmic}
\end{algorithm}

We call our procedure \textsc{SLoPP} (structural learning of PSDDs under PCWA). Algorithm \ref{alg:slopp} depicts the \textsc{SLoPP} workflow. This is a recursive procedure, that starts growing the circuit from the top. A clustering algorithm (line 11) is executed on the columns of the database corresponding to the left variables of the vtree. The clusters obtained in this way are used to generate the primes (line 13), while the subs are made of the records for the right variables induced by the same clusters (line 14). The probabilities assigned to these product units are proportional to the cardinality of the corresponding clusters (line 16). When we finally obtain databases over single variables (lines 2-8) univariate mass functions or simple literals are specified depending on the frequencies of the two Boolean states in the column. The only parameters of the algorithm are the number of clusters to be returned (or the criteria to select this number) and the minimum amount of records required to run the clustering algorithm. If the number of records is smaller than this threshold, a single cluster with all the records is returned.

As discussed in Section \ref{sec:background}, given a database $\mathcal{D}$, CWA allows to describe the possible states of the affairs induced by the database as a DNF $\phi(\mathcal{D})$.
The PSDD we learn represents a \emph{relaxation} of $\phi(\mathcal{D})$, i.e,  $\phi(\mathcal{D})$ logically implies the formula encoded by the learned PSDD.  This is formalised by the following result, whose derivation is detailed in the appendix.

\begin{prp}\label{prop:slopp} Algorithm \ref{alg:slopp} returns a valid PSDD representing a relaxation of the formula induced by the input database, i.e., 
$\phi(\mathcal{D})$ logically implies  $\interp{{\textsc{SLoPP}}(\mathcal{D},v)}$.
\end{prp}

The aforementioned relaxation happens because \textsc{SLoPP} applies a PCWA at each recursive call, as we shall explain with a simple example. Consider Table \ref{tab:data} and vtree $v$ in Figure \ref{fig:vtree1} whose left variables are $\bm{v^l}=\{X_1, X_2\}$ and whose right variables are $\bm{v}^r=\{X_3,X_4\}$. The horizontal lines in that table separate the output of a clustering of the records in $\mathcal{D}^{\bm{v}^l}$, these inducing the primes of the root node, whose corresponding partition $\mathcal{D}^{\bm{v}^r}$ induces its subs. For instance, the second element is induced by the records corresponding to the third, fourth and fifth line in Table \ref{tab:data}. In particular prime $p_2$ is  induced by the sub-table described by the third, fourth and fifth lines restricted to the first two columns (associated to $X_1$ and $X_2$), whereas the restriction of those three lines to the last two columns (associated to $X_3$ and $X_4$) induces the corresponding sub $s_2$. By combining the records of these two sub-tables, we see that three new \emph{virtual records}, $[1,0,1,1], [1,0,1,0]$ and $[0,1,0,1]$ arise. Those three virtual records have zero counts, as showed at the bottom of Table \ref{tab:data}.  
This notwithstanding, they are considered to be possible by the PSDD returned by \textsc{SLoPP}. Indeed, this PSDD, depicted in Figure \ref{fig:psdd}, induces a mass function that assigns non-zero probability exactly to all the instances appearing in $\mathcal{D}$, included the virtual records. Stated otherwise, these records coincide with the joint states satisfying the Boolean formula encoded by the PSDD that is returned by \textsc{SLoPP} when applied to Table \ref{tab:data}. 



\section{Preliminary Experiments}\label{sec:ex}
A first implementation of Algorithm \ref{alg:slopp} has been achieved within the Juice\footnote{\href{https://github.com/Juice-jl}{\tt github.com/Juice-jl}.} (Julia) library for PCs \citep{DangAAAI21}. The code, freely available as a simple Julia notebook,\footnote{\href{https://github.com/IDSIA-papers/2021-TPM}{\tt github.com/IDSIA-papers/2021-TPM}.} is highly experimental and not optimized for fast performance. For this reason, in this position paper, we discuss the results of a very preliminary validation based only on two small databases as available in Juice and whose features are detailed in Table \ref{tab:db}. A deeper analysis based on a larger benchmark is a necessary future work.

\begin{table}[htp!]
\centering
\begin{tabular}{lrrr}
\toprule
Name&$|\mathcal{D}_{\mathrm{train}}|$&$|\mathcal{D}_{\mathrm{test}}|$&$|\bm{X}|$\\
\midrule
Nltcs&16181&3236&16\\
Plants&17412&3482&69\\
\bottomrule
\end{tabular}
\caption{Characteristics of the benchmark databases.}
\label{tab:db}
\end{table}

Given a complete database of Boolean variables, \textsc{SLoPP} needs a vtree over those variables. Here we adopt the state-of-the-art techniques proposed by \cite{DangAAAI21} and based on Chow-Liu trees as included in the (Juice implementation of the) \textsc{Strudel} algorithm. The PSDD returned by this algorithm is used to evaluate the performance of our method, together with a fully-factorised model intended to provide a trivial baseline level.

To cluster records we use the k-means algorithm. Here we set the number of clusters $k$ constant (and equal to two and three), while a threshold on the database size $d$ to create multiple clusters is also specified (and if this is not the case a single cluster with all the data is used).

As a consequence of Proposition \ref{prop:slopp}, the PSDD returned by \textsc{SLoPP} might be consistent even with records not present in the training set. Yet, it might be possible that testing records not available in the training set would be inconsistent with the PSDD. We discard those records from the test database and we denote as $\gamma$ their number. This clearly gives an unfair advantage to \textsc{SLoPP} if compared with 
\textsc{Strudel} or with the fully-factorised model, as both these PSDDs encode the tautology (i.e., no KB). Yet, for such a preliminary study, we are interested in evaluating the strength of such advantage in terms of performance or its relevance as shown by $\gamma$. 
Tables \ref{tab:ll} and \ref{tab:ll2} depict the test log-likelihood on the consistent records for different values of the parameters.


\begin{table}[htp!]
\centering
\begin{tabular}{lrrrrrr}
\toprule
$\mathcal{C}$&$k$&$d$&$\gamma$&$LL_{\mathcal{C}}$&$|\mathcal{C}|$\\
\midrule
\textsc{Fully Fact.}&-&-&-&-35'844&79\\
\textsc{Strudel}&-&-&-&-20'054&786\\
\textsc{SLoPP}&2&20&4&-22'148&1'229\\
\textsc{SLoPP}&2&50&7&-23'441&1'232\\
\textsc{SLoPP}&3&20&23&{\bf -19'744}&2'258\\
\textsc{SLoPP}&3&50&15&-20'176&2'033\\
\bottomrule
\end{tabular}
\caption{Test log-likelihood with Nltcs for \textsc{SLoPP}, \textsc{Strudel} and fully factorised models. Different experiments with different number of clusters $k$ and threshold on the minimum cluster size $d$ are reported. Parameter $\gamma$ is the number of inconsistent test instances, while $|\mathcal{C}|$ is the number of nodes of the circuit $\mathcal{C}$ obtained with the different approaches.}
\label{tab:ll}
\end{table}



\begin{table}[htp!]
\centering
\begin{tabular}{lrrrrrr}
\toprule
$\mathcal{C}$&$k$&$d$&$\gamma$&$LL_{\mathcal{C}}$&$|\mathcal{C}|$\\
\midrule
\textsc{Fully Fact.}&-&-&-&-133'389&344\\
\textsc{Strudel}&-&-&-&-38'266&2'503\\
\textsc{SLoPP}&2&20&582&-40'442&71'602\\
\textsc{SLoPP}&2&50&594&-39'354&69'529\\
\textsc{SLoPP}&3&20&713&-48'329&103'742\\
\textsc{SLoPP}&3&50&793&{\bf -36'003}&95'889\\
\bottomrule
\end{tabular}
\caption{Test log-likelihood with Plants. The same setup as in Table \ref{tab:ll} is considered.}
\label{tab:ll2}
\end{table}


As a positive remark we notice comparable performance levels between \textsc{SLoPP} and \textsc{Strudel} and a relatively low number of inconsistent test records. The most critical point seems to be the huge size of the PSDDs returned by \textsc{SLoPP}: for Plants there are more than an order of magnitude larger than the ones produced by \textsc{Strudel}. This is probably related to the fact that the current implementation of our algorithm is only coping with singly-connected topologies, thus creating unnecessarily large circuits. Forcing the creation of multiple connections in the circuit could lead to smaller circuits.



\section{Conclusions}\label{sec:conc}
A new structural learning algorithm for PSDDs has been presented. The algorithm uses the input data as a knowledge base to be relaxed by a PCWA approach. A very preliminary validation suggests that the algorithm might lead to reliable models to be used in frameworks where (some form of) PCWA is a tenable hypothesis. Besides an extensive experimental analysis, as a future work we intend to embed the learning of the vtree within the \textsc{SLoPP} architecture in order to better exploit the statistical information in the data set. In order to keep under control the size of the circuit returned by the algorithm, the dynamic generation of multiple connections when more elements involving the same sub-circuit are present should be also considered. A relaxed version of this idea might be used to obtain even smaller models. An integration between \textsc{SLoPP} and \textsc{LearnPSDD} (e.g., achieved by using the output of the first algorithm as an input for the second) might also be considered to train better models. Finally, belief revision techniques could be considered to solve the training-testing-set inconsistency issue discussed in the experimental section.


\section*{Proofs}
\paragraph{Proof of Proposition \ref{prop:slopp}}
It is an easy exercise to verify that the output of \textsc{SLoPP} is a valid PSDD.

To see that the formula encoded by   $\textsc{SLoPP}(\mathcal{D},v)$ is a relaxation of $\phi(\mathcal{D})$, we proceed by induction on the number $n$ of variables in $\mathcal{D}$. If $n=1$, $v$ is a leaf and $\textsc{SLoPP}(\mathcal{D},v)$ is a trivial PSDD consisting in a single unit $X$, $\neg X$ or $(\top, \tfrac{|\mathcal{D}^{X=1}|}{|\mathcal{D}|})$. In each case the encoded formula is by definition $\phi(\mathcal{D})$. Assume that the statement holds for a number of variables strictly smaller than a fixed $n>1$. Then, for a database $\mathcal{D}$ over $n$ variables and associated vtree $v$, consider $\textsc{SLoPP}(\mathcal{D},v) = \{(p_i,s_i,\theta_i) \}_{i=1}^k$. For $1\leq i \leq k$, $p_i = \textsc{SLoPP}(\mathcal{D}^{\bm{v^l}}_{J_i}, v^l)$ and $s_i=\textsc{SLoPP}(\mathcal{D}^{\bm{v^r}}_{J_i}, v^r)$, where $J_1, \dots, J_k$ are the clusters obtained from $\mathcal{D}^{\bm{v^l}}$. 

Moreover, $\interp{\textsc{SLoPP}(\mathcal{D},v)}= \mathop{\bigvee}_{i=1}^k \interp{p_i}\wedge \interp{s_i}$, and by induction hypothesis $\interp{p_i}$ and $\interp{s_i}$ are relaxations of $\phi(\mathcal{D}^{\bm{v^l}}_{J_i})$ and $\phi(\mathcal{D}^{\bm{v^r}}_{J_i})$, respectively. 

This means that
$\interp{p_i}$ implies  $\phi(\mathcal{D}^{\bm{v^l}}_{J_i})$ and $\interp{s_i}$ implies $\phi(\mathcal{D}^{\bm{v^r}}_{J_i})$, and thus we get that $\interp{p_i}\wedge \interp{s_i}$ implies their conjunction. We can immediately conclude that
$\interp{\slopp(\mathcal{D},v)}$ implies $\mathop{\bigvee}_{i=1}^k \phi(\mathcal{D}^{\bm{v^l}}_{J_i}) \wedge \phi(\mathcal{D}^{\bm{v^r}}_{J_i})$.

Now, for $1\leq i\leq k$, both $\phi(\mathcal{D}^{\bm{v^l}}_{J_i})$ and $\phi(\mathcal{D}^{\bm{v^r}}_{J_i})$ are DNFs whose conjunctive clauses are induced by the rows of $\mathcal{D}^{\bm{v^l}}_{J_i}$ and of $\mathcal{D}^{\bm{v^r}}_{J_i}$, respectively. By taking their conjunction we combine their clauses and clearly get, among others, $ \phi(\bm{x}^i) $, the conjunctive clause induced by the  $i^{th}$ record of $\mathcal{D}$. This yields the desired conclusion.

\bibliography{biblio}
\end{document}